\documentclass[10pt,twocolumn,letterpaper, times]{article}
\usepackage[margin=25mm]{geometry}

\usepackage{graphicx}
\usepackage{amsmath}
\usepackage{amssymb}
\usepackage{booktabs}

\usepackage{times}
\usepackage{epsfig}
\usepackage{graphicx}
\usepackage{xcolor}
\usepackage{caption}
\usepackage{subcaption}
\usepackage{array}

\def\abstract
   {
   \centerline{\large\bf Abstract}%
   \vspace*{12pt}%
   \it%
   }

\usepackage{biblatex} %Imports biblatex package
\makeatletter
\newcommand\extralabel[2]{{\edef\@currentlabel{\@currentlabel#2}\label{#1}}}
\makeatother

\usepackage[breaklinks,colorlinks]{hyperref}

\usepackage[capitalize]{cleveref}

\title{\textbf{Photometric Correction for Infrared Sensors}}

\author{Jincheng Zhang$^1$, Kevin Brink$^2$, Andrew R. Willis$^1$\\
$^1$University of North Carolina at Charlotte, $^2$Air Force Research Laboratory\\
$^1$9201 University City Blvd. Charlotte, NC 28223\\
$^2$Eglin AFB, FL, 32542 \\
{\tt\small $^1$\{jzhang72, arwills\}@uncc.edu, $^2$kevin.brink@us.af.mil}
}
\date{} % Comment this line to show today's date

\begin{document}
\maketitle

%%%%%%%%% ABSTRACT
\begin{abstract}
  Infrared thermography has been widely used in several domains to capture and measure temperature distributions across surfaces and objects. This methodology can be further expanded to 3D applications if the spatial distribution of the temperature distribution is available. Structure from Motion (SfM) is a photometric range imaging technique that makes it possible to obtain 3D renderings from a cloud of 2D images. To explore the possibility of 3D reconstruction via SfM from infrared images, this article proposes a photometric correction model for infrared sensors based on temperature constancy. Photometric correction is accomplished by estimating the scene irradiance as the values from the solution to a differential equation for microbolometer pixel excitation with unknown coefficients and initial conditions. The model was integrated into an SfM framework and experimental evaluations demonstrate the contribution of the photometric correction for improving the estimates of both the camera motion and the scene structure. Further, experiments show that the reconstruction quality from the corrected infrared imagery achieves performance on par with state-of-the-art reconstruction using RGB sensors.
\end{abstract}

%%%%%%%%% BODY TEXT
%\vspace{-0.5cm}

\section{Introduction}

% Establish the topic area
% Introduction to the IR sensors and their application: https://www.sciencedirect.com/science/article/pii/S1350449521001262?via%3Dihub

Infrared (IR) sensors have essential applications in a wide variety of sensing contexts and have recently seen much interest as a sensor to facilitate autonomous ground and aerial vehicle navigation. Heightened interest is largely due to the IR sensor's capability to sense accurate scene image data under low light conditions. This is particularly useful in contexts where illumination is unavailable, e.g., navigating at night and when the visible light sources, surface appearance textures, and surface reflections introduce difficulties to visible light algorithms, e.g., camouflaged targets \cite{grimming2021lwir, pinchon2018all, rankin2011unmanned, kim2017infrared}. 

Infrared image sensors consist of a grid of radiation-sensitive optoelectronic components that are sensitive to radiated energy having wavelengths from the IR frequency band which spans wavelengths $\lambda \in [0.7\mu m, 1000 \mu m]$. The IR frequency band is commonly divided into the NIR (Near IR, $\lambda \in [0.7\mu m, 2.5 \mu m]$), MWIR (Mid-Wavelength IR, $\lambda \in [2.5\mu m, 5 \mu m]$), LWIR (Long-Wavelength IR, $\lambda \in [8\mu m, 15\mu m]$), and the FIR (Far IR, $\lambda \in [15\mu m, 1000 \mu m]$). 

%Technology improvements in infrared imaging have facilitated the use of thermal imaging in an increasing number of fields, especially in the field of computer vision. % Researchers have applied infrared images to deep learning to perform different computer vision tasks, for example, object detection and tracking \cite{}, medical research, and image 
%Most of these methods only employ 2D thermal images since they are easily obtainable. However, even though the thermal images can provide heat distributions, they do not provide depth or 3D information about the heat source. For example, it is possible to deduce from a 2D thermal image that a heat source exists, but the distance between the heat source and the thermal scanner is unknown from a single image ~\cite{Schonberger_2016_CVPR}.

This article proposes a new photometric correction for thermographic infrared cameras. Thermographic imaging devices use microbolometer sensors to detect radiation in the MWIR and LWIR frequency ranges. For simplicity, we will refer to microbolometer sensors simply as IR sensors. \emph{It is important to note that other sensing devices are required to capture images in the NIR and FIR wavelengths and the results of this article do not apply to these IR sensor types \cite{hyll2012infrared}.}

Similar to related work for RGB cameras \cite{engel2016photometrically}, photometric correction for these IR image sensors promises to improve the accuracy of measured pixels by estimating the underlying scene irradiance responsible for generating a pixel value. Accurate estimates of the scene irradiance promise to improve the output quality of many existing computer vision algorithms \cite{suhr2009kanade, beauchemin1995computation}. This is especially true for the large class of algorithms that assume scene points are ``viewpoint invariant," i.e., the Lambertian assumption, for modest changes in the camera pose. 

The proposed photometric correction is tailored to microbolometer pixel sensors whose characteristic response is known in the literature \cite{boudou2019ulis, kohin2004performance} and distinct from the response of RGB image sensors. Photometric correction for these image sensors promises to allow vision algorithms developed for popular RGB image sensors to translate to infrared sensors better. In fact, our results indicate that some aspects of the estimates provided by infrared sensors may be more accurate than corresponding estimates from RGB sensors (see Section \ref{sec:exp}).

The contributions of this article include novel theoretical and experimental results. These results include:

\begin{itemize}
    \item A novel photometric correction model for infrared (microbolometer) sensors is proposed. 
    \item A novel application of this model to SfM problems and a comparison of SfM accuracy with IR vs. RGB sensors.
    \item Experimental work showing the proposed photometric correction model can improve algorithm performance for SfM problems.
    \item Experimental work showing the proposed photometric correction model can provide performance benefits over traditional RGB sensors for SfM problems. 
\end{itemize} 

To our knowledge, this work constitutes the first quantitative analysis of SfM for IR sensors in the literature. The impact of this photometric correction is important to several sensor types, sensing conditions, and sensing contexts which include: (1) uncooled IR sensors, (2) IR sensors that move or observe moving scenes, (3) IR sensors that operate in high-temperature environments and (4) IR sensors that operate at high frame rates. Each of these circumstances can lead to circumstances where the value of sensed pixels from prior image frames can significantly offset the value of subsequent measurements leading to ghosting effects of the moving object in the video \cite{FLIR_boson_tau2}. The proposed IR photometric correction approach seeks to compensate for these effects.

%\kb{Your 2nd bullet is very different from your 3rd and 4th, are your contributions what you did (created, implemented, and tested a correction model) or are they the results of the tests? I think its the former (and you can sprinkle the fact it's better in there as you call the ``what'' out in your bullets}

%\jz{(The model is more applicable for uncooled infrared sensors.)}
%We show the possibility of reconstructing a 3D scene from infrared images and provide analysis to demonstrate that with proper calibration of the sensor, the reconstruction quality from infrared data can be improved. Initial results show that such improved quality is comparable to even better than one from RGB data, especially in a situation where illumination condition is not ideal, for example, when the data is captured at night. To the best of our knowledge, our work is the pioneer which explores the applicability of infrared data used in structure from motion tasks.

%\aw{Any context with high framerate and moving scene pixels. Transitions between no excitation and saturation (values 0 and 255 for 8-bit sensors) need photometric compensation. These circumstances maximize the error of heating and cooling.}

\section{Related work}
%\aw{Paragraph here: You state the focus of the article is to compensate for pixel value variability associated with the physics of uncooled microbolometers. The article then applies this model for the purpose of improving upon state-of-the-art Structure from Motion solutions using these sensor types.}
The contribution of this work is to propose and apply a photometric correction algorithm specific to IR image sensors. Other than \cite{das2021online} where a linear sensor response model is used to calibrate IR sensors, we have been unable to find other references within the computer vision literature detailing similar approaches. For this reason, we use the related work section to discuss research on models to characterize the response microbolometer image sensor pixel values. Experimental work integrates the proposed photometric correction to a state-of-the-art SfM algorithm and a component of the literature review discusses SfM algorithms and how advances in photometric correction for RGB sensors have served to improve SfM estimates.

\subsection{Infrared Imaging with Microbolometers}
%\aw{Main message: Show the basics of how IR pixels respond to IR radiation. Discuss what happens when the sensor is unable to cool completely. This circumstance can arise when there is a short duration between measurements when the sensor is in a hot environment, and when the pixel value is near saturation.} 

An infrared camera is a device that converts infrared radiation into a visual image that depicts temperature variations across an object or scene. Infrared radiation is a characteristic of all objects that have a temperature higher than absolute zero (zero Kelvin or -273 Celsius) \cite{WILSON2021202}. Thermal energy radiated by scene objects is focused onto the sensor image plane where a grid of microbolometer sensors is placed to convert the optical energy focused onto each grid element into a pixel voltage indicative of the object temperature. 

The physical response of each pixel element is governed by a heating and cooling mechanism as the camera shutter is opened and closed. Similar to RGB optical sensors, microbolometer sensors integrate incident radiation into stored charge when the camera shutter is open and dissipate the stored charge when the camera shutter is closed. This process generates a response analogous to an RC circuit driven by a square wave excitation where the square wave period is determined by the exposure time and its amplitude is determined by the radiated energy of the scene onto the pixel sensor. The rate of integration and dissipation is driven by a sensor-specific time constant, $\tau$.

One shortcoming of microbolometer sensors is their response time. RGB pixel sensors based on CMOS technology have been shown to have time constants $\tau \in [1\mu s, 500 \mu s]$ with the median sensor performance across RGB sensors $\tau \approx 10 \mu s$ at room temperature \cite{7731134}. Typical response times for microbolometer sensors are $\tau \in [8ms, 15ms]$ which is more than two orders of magnitude larger. Long response times lead to ``ghosting'' of sensed values where the value of a pixel in a past frame persists in the current frame creating a spatio-temporal blur of moving objects in sensed IR images.
%\jz{Such conversion leads to alterations in the temperature of the pixel, which consequently modifies the electrical resistance of the thin film resistor of that pixel. Afterward, the readout integrated circuit transmits a voltage to every microbolometer component, and the signal proportional to the heat absorbed by each detector forms the foundation of a video image \cite{understandinfrared}.}

%Uncooled microbolometers integrated signal from an earlier frame \cite{FLIR_boson_tau2}. As a result, sensed microbolometer pixel values have that was collected in previous frames. This causes the value of sensed pixels from prior frames to offset the value of subsequent measurements. This circumstance can arise when there is a short duration between measurements, for example, the sensor is taking on fast-moving targets, or when there is an extremely hot object to be measured and the pixel value is near saturation. This characteristic of the microbolometer prevents the infrared sensor from accurately measuring the infrared radiation of objects when the camera is moving with respect to the scene. In order to overcome the drawback of the uncooled infrared sensors (microbolometers) when used for SfM, the proposed IR photometric correction approach seeks to compensate for the heat residual from the frame history.

\subsection{Structure from Motion}

Structure from Motion is the process of reconstructing a 3D structure from its projections into a series of images taken from different viewpoints. It commonly starts with feature extraction and matching, followed by geometric verification \cite{schonberger2016structure}, where the feature matching searches for feature correspondences by finding the most similar feature between image pairs with scene overlapping and comparing the similarities between features. This requires the detected feature to be accurate and reliable throughout the image sequence. We note work in the literature on the related problem of visual odometry from IR image data which estimates the ``M" or camera motion estimation component of the SfM problem \cite{mouats2015thermal, khattak2019keyframe, khattak2020keyframe}. Work in \cite{emilsson2014chameleon} demonstrates that RGB SfM algorithms may be used on IR cameras for use in firefighting scenarios with some success.

The photometric correction has proven to be important for improving the result of SfM estimates \cite{engel2017direct,semi-direct, largescale}. Direct Sparse Odometry (DSO) \cite{engel2017direct} achieves state-of-the-art performance by taking full advantage of photometric camera calibration, including lens attenuation, gamma correction \cite{engel2016photometrically}, and brightness correction. The lens attenuation and gamma correction both require prior knowledge about the sensor and lens being used for solving SfM problems, the brightness correction, instead, is an empirical model that takes into account automatic exposure changes and compensates the pixel values in order to make them more consistent and stable across a sequence of images. The brightness correction function in DSO is given by $e^{-a}(I-b)$ where $a$, $b$ are the correction parameters, and $I$ is the pixel value. In their work, the authors show that a naive brightness constancy assumption used in other approaches like LSD-SLAM \cite{engel2014lsd} or SVO \cite{forster2014svo} significantly decreases the SfM accuracy. This article builds on these concepts by translating this intuition from CMOS-based RGB image sensors to microbolometer-based IR image sensors. A photometric correction model for microbolometer pixel values is proposed in order to improve the performance for SfM problems using IR data.

%-------------------------------------------------------------------------
\section{Methodology}

Photometric correction is necessary for uncooled infrared image sensors because the microbolometers used by these sensors have a response that is entirely different from photon detector imagers such as RGB cameras. Specifically, uncooled microbolometer devices require a certain portion of each frame time to integrate signals. Hence for high-speed temperature measurement with short frame time, pixels in a microbolometer usually do not have enough time to reach the temperature of the scene to be measured (a steady temperature state) before the pixels receive new radiance from objects in the next frame. Moreover, microbolometers do not have a mechanism to reset the integrated signal from the previous frame, and therefore signals captured in previous frames will have a residual impact on the microbolometer pixel reading in the current frame. Both of these factors result in an ``inaccuracy'' of sorts in pixel values in the images generated by infrared sensors as they are not determined solely by the ``current scene'' and this results in unreliable reconstruction results via structure from motion.

\begin{figure}
     \centering
     \begin{subfigure}[b]{0.22\textwidth}
         \centering
         \includegraphics[width=\textwidth]{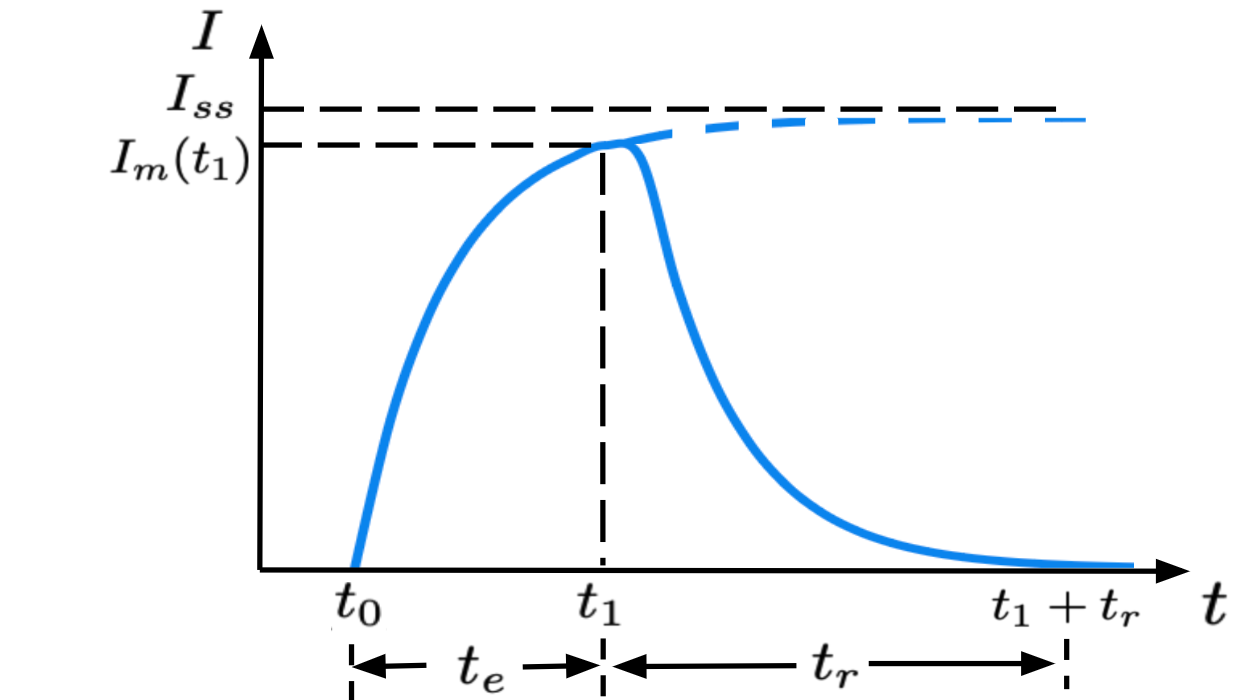}
         \caption{Microbolometers heating and cooling process}
         %\label{fig:heating_and_cooling}
     \end{subfigure}
     \quad
     \begin{subfigure}[b]{0.22\textwidth}
         \centering
         \includegraphics[width=\textwidth]{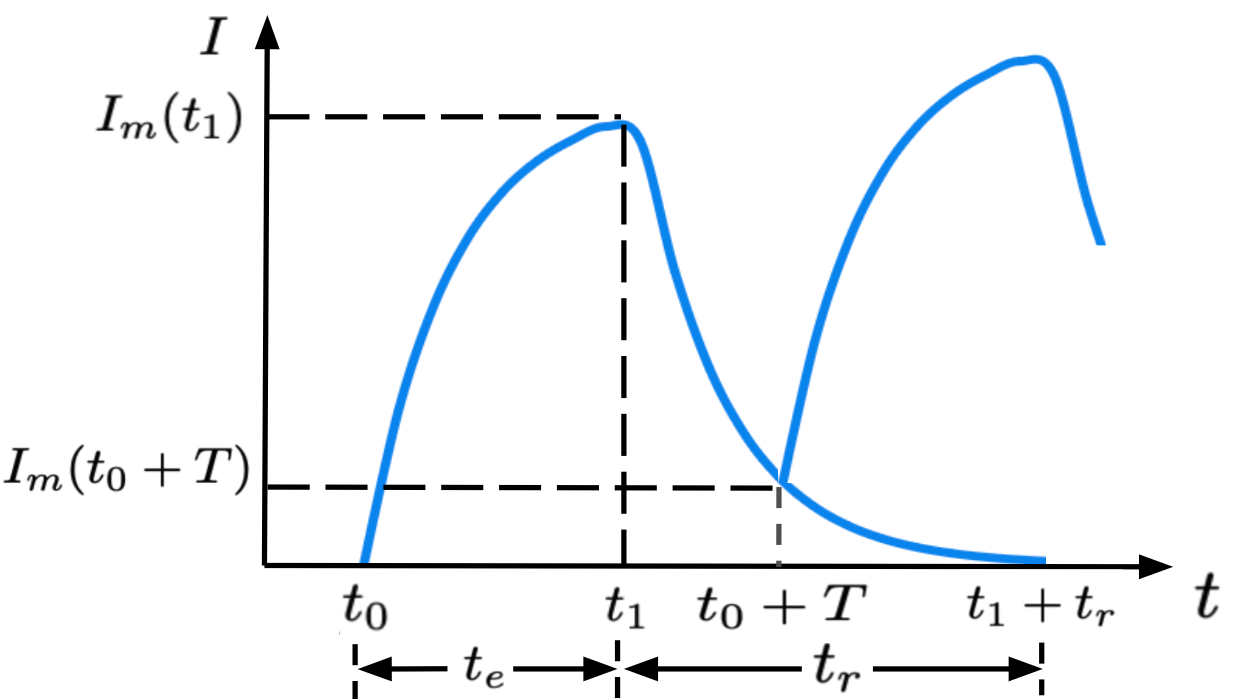}
         \caption{Microbolometers behavior for two consecutive frames}
         %\label{fig:combine}
     \end{subfigure}
     \caption{An illustration of microbolometer pixels' behavior during the heating and cooling process. (a) Pixels do not have enough time to reach the temperature of the scene being measured. (b) In two (or more) consecutive frames the later frame(s) has a memory of the energy residual in the previous frame(s).}
     \label{fig:models}
     \extralabel{fig:models:a}{(a)}
     \extralabel{fig:models:b}{(b)}
\end{figure}

The proposed model for photometric correction of microbolometer measurements consists of two parts: (1) a heating model which characterizes the IR pixel response during a frame exposure and (2) a cooling model which characterizes the IR pixel response when the sensor is not exposed. This section describes these models using continuous differential equations and combines these models into a comprehensive photometric correction model for IR pixels measured at arbitrary framerates. We discuss camera models that allow these corrections to be position invariant and, under these circumstances, algorithms can quickly apply photometric correction across all image pixels using lookup tables to improve their performance. We then integrate this model into the DSO SfM solution and detail how the SfM problem is modified by the integration of this new photometric correction.

\subsection{Model for Pixel Heating}
%\aw{Where's the diff. eq? Write it, solve it, and then use your solution. We find the value of the forced response at a steady state due to the excitation of the pixel from the associated portion of the 3D scene.}

%Each pixel in an infrared sensor physically heats up during the integration time (exposure time) when measuring the temperature of objects. However, this does not mean the sensor's pixels can be read out within the same time period \cite{FLIR_app_story}. Sensor's pixels require more time than the integration time to reach a steady state to accurately reflect the temperature of the measured objects. \kb{It is okay to repeat things, but you need to have them at a very low fidelity all but one time - you either allude to details you'll provide later, or you remind of details you provided further. Make sure you don't explain the same concept in medium detail too often, especially in back-to-back subsections. }

%\aw{heating section}
Heating the IR pixel is modeled as a differential equation with a unit step forcing function, $\mu(t)$, where the amplitude of the step is proportional to the scene irradiance. We then seek to calculate the steady pixel value that reflects the unknown irradiance of the scene point which corresponds to the asymptotic of the step response. The rate of the convergence of the pixel value to the steady state response is determined by the \emph{heating time constant}, $\tau_{h}$. The process of pixel heating up can be depicted in Fig. \ref{fig:models:a}.

The model denotes the initial measurement time as $t=t_{0}$ and uses the ``black body" assumption to set the initial value of the IR pixel, i.e., $I_{m}(t_{0}) = 0$.

%\aw{Let $t_{e}=t_{1}-t_{0}$ denote the exposure time. Let $\gamma=\frac{T}{t_{1}-t_{0}}$ denote the ratio of the frame duration to the exposure time. If $\gamma=0.5$ then the heating and cooling effect have equal time.} 

%\aw{FLIR documentation to show $\tau$ is different for heating and cooling process.}

%\kb{You are saying we and our a lot. I am all for informality vs. overly dry or long text to avoid we, etc. But if you can easily avoid it, I think you may want to try (although some journals, etc. are much less formal so I might be wrong).  Our model can be the model. We do X by, X is done by, etc.  But please use we and our if it's not a simple switch. }

%Inspired by other research in image sensors \cite{FLIR_app_story, york2011fundamentals}, we characterize the heating of an IR pixel with a first-order differential equation

A first-order differential equation model is provided in Eq. \ref{eq:heating_correction_eq} to characterize heating an IR pixel.

%\jz{simplify the equation; make them a function of t; read out -> measured pixel value; make an analogy to RL circuit?}

\begin{equation}\label{eq:heating_correction_eq}
\begin{gathered}
\tau_{h} \frac{\sl{d}I_{m}(t)}{\sl{d}t} + I_{m}(t) = I_{ss} \\
I_{m}(t_{0}) = 0
\end{gathered}
\end{equation}
where $I_m(t)$ is the pixel intensity value measured by the camera sensor at time $t$, and $I_{ss}$ is the steady state pixel intensity value if the microbolometers were given sufficient heating time. 

Let $t_{e}=t_{1}-t_{0}$ denote the exposure time, meaning that the shutter is open from the beginning time $t_0$ to time $t_1$. Solving the first-order differential equation at $t=t_1$ gives 

\begin{equation}\label{eq:heating_correction_sol}
I_{ss}=\frac{I_{m}(t_0+t_e)}{{(1-e^{-\frac{t_e}{\tau_h}})}}
\end{equation}

By modeling the heating process of a microbolometer, we find the value of the forced response at a steady state due to the excitation of the pixel from the associated portion of the 3D scene.

\subsection{Model for Pixel Cooling}
%Unlike typical rolling shutter cameras, uncooled infrared cameras do not have a mechanism to reset the integrated signal from the previous frame. As a result, a microbolometer pixel will have a memory of the signal that was collected in previous frames. This characteristic of the camera causes image artifacts such as tails on hot moving objects \cite{FLIR_boson_tau2}. \kb{same comment - make sure you're only detailed once}

%\aw{cooling section}
Cooling the IR pixel is modeled as the natural response of the same differential equation after the heating forcing function, $\mu(t)$, has been set to zero. We then seek to calculate the measured pixel value at the time $t=t_{0}+T$ where $T$ is the time of each frame. This pixel value will then contribute as a non-zero initial condition for the subsequent image at time $t_{0}+T$. The decay rate of the pixel value is determined by the \emph{cooling time constant}, $\tau_{c}$. The process of pixel cooling is depicted in Fig. \ref{fig:models:a}. 

The model denotes the excitation of the pixel at the time that the forcing function is removed as $t=t_{1}$ when the shutter is closed, and uses the value of the measured pixel at $t=t_{1}$ as the initial value of the IR pixel, i.e., $I_{m}(t_{1}) = I_{0}.$

Similarly, another first-order approximation is used to describe the cooling process.

\begin{equation}\label{eq:cooling_correction_eq}
\begin{gathered}
{\tau_c}\frac{\sl{d}I_{m}(t)}{\sl{d}t} + I_{m}(t) = 0 \\
I_{m}(t_1) = I_{0}
\end{gathered}
\end{equation}

Let $t_r=t_0+T-t_1$ denote the sensor readout time when the shutter is closed. Solving this first-order differential equation at time $t=t_0+T$ gives 

\begin{equation}\label{eq:cooling_correction_sol}
    I_{m}(t_0+T)=I_{0} e^{-\frac{t_r}{\tau_c}}
\end{equation}

By modeling the cooling process of a microbolometer, we find the pixel value at the end of each frame period, which is also the initial measurement for the next frame.

\subsection{Complete Video Sensing Model \label{sec:ir_sensor_model}}

We consider sensors that record images at a framerate of $f_s$ or equivalently having a temporal sample period $T=\frac{1}{f_s}$. The time interval between each frame, $T$, is further subdivided into a measurement or exposure time during which time the shutter is open, $t_{e}$, and a readout time during which the shutter is closed, $t_{r}$. During the exposure time period, the microbolometer is heated. During the period that the sensor reads out the pixel values, the microbolometer is cooling. Fig. \ref{fig:models:b} shows the pixel value convergence when the microbolometer is heating and the heat residual from the previous frames left on the new frame when the microbolometer is cooling. We seek to calculate the steady state pixel value excited by a scene point with the prior heat residual removed. Assuming that the effect from previous frames is dominated by the most recent prior frame when there are no drastic temperature gradients in the scenes, %\kb{when is this valid - maybe make a disclaimer - ``an assumption, but a reasonable one for scenes without drastic temperature gradients" or something?}, 
the earlier frames are ignored in the model. The complete video sensing model in Eq. \ref{eq:combined_correction} merges these two models into a comprehensive model for the pixel response, $I_i'(t_{e},t_{r})$ at frame $i$ while recording a time-sequence of images.

\begin{equation}
    I_i'(t_{e},t_{r})=\frac{I_i-I_{i-1}(e^{-\frac{t_{r, i-1}}{\tau_c}})}{(1-e^{-\frac{t_{e,i}}{\tau_h}})}
\label{eq:combined_correction}
\end{equation}
where $I_{i-1}$ and $I_i$ are two continuous frame, $t_{r,i-1}$ is the readout time in the previous frame $i-1$, and $t_{e,i}$ is the exposure time of the current frame $i$.

By combining the heating model and cooling model, the irradiance, or the temperature of the measured scene point, can be more accurately reflected by the stable pixel value $I'_i$.

In the remainder of this paper, $I_i$ will always refer to the photometrically corrected image $I'_i$, except where otherwise stated.

%\aw{Solve for the true scene irradiance. $B_{i} = I_{ss}$}

%\subsection{Position Invariance}

%\aw{Pixel correction is position invariant when lens effects are properly compensated for \textbf{Is this in the results? If so indicate that you will show this} Explain that this is true because all pixels use the same microbolometer sensing elements.}

%\aw{this discussion applies across all imaging algorithms, Optical flow, stereo reconstruction, odometry, and Structure from Motion.}

%\jz{as long as we compensate for lens effects, this model can be applied across all locations in the image using a look-up table and it will improve the accuracy of the measurement to make it more correctly reflect. It is fast.}

%Ideally, the sensitivity of image acquisition and digitization devices should be position invariant in the image. This assumption, however, is not valid in practice due to factors such as the optical properties of camera lenses. Such degradation can significantly impair the structure from motion algorithm that relies heavily on the stability of the pixel intensity data. To reduce the lens effects, a vignetting correction is applied before the use of the proposed photometric correction. This allows the accuracy improvement of the measurement and enables the photometric correction later to be position invariant. With the lens effects compensated properly, the performance of the photometric correction will be more stable and more efficient since the position-invariant correction can be applied across all locations in the image using a look-up table.  

\subsection{IR Sensor-Based Structure From Motion (SfM)}

%\aw{Briefly describe DSO as in the related introduction paragraph. Stress the contribution of DSO as a state-of-the-art Structure from Motion (SfM) method. In [add cites] the authors develop advanced photometric correction models for both camera lens and RGB pixel sensing compensation. This is coupled with camera calibration data to perform highly-accurate SfM at real-time rates with impressive 3D structure reconstruction results.} 

%\aw{Remind the reader of the photometric correction model in the related work (reference). Then indicate how your photometric model is substituted into DSO.}

%\aw{Clearly state how the DSO SfM algorithm was changed to implement SfM for infra-red pixel sensors.} 

In DSO the authors develop advanced photometric correction models for both camera lens and RGB pixel sensing compensation. This is coupled with camera calibration data to perform highly-accurate SfM at real-time rates with impressive 3D structure reconstruction results. To leverage such a system and apply it to infrared sensors, the brightness transfer model used for RGB sensors in DSO is replaced by our infrared sensor photometric correction model with the following modifications in the SfM algorithm.

\begin{itemize}
    \item Added a time history (previous sensed values) to tracked pixels.
    \item Modified the optimization approach to use the derivatives and Hessian of our photometric correction model. %\aw{maybe you write these down as equations?}
    %\item Modified the optimization source/target to only perform the forward correction.
\end{itemize}

To reconstruct a 3D scene from infrared images using structure from motion, a map is computed to associate two pixels in different frames that both correspond to the same 3D scene point. The photometric error between them is defined in a similar way as \cite{engel2017direct}. For a point, $\mathbf{p}$ in reference frame $I_i$, observed in target frame $I_j$, the photometric error, given by Eq. \ref{eq:model_formulation}, is formulated as the weighted Sum of Squared Differences (SSD) over a small neighborhood of pixels.

\begin{equation}
\label{eq:model_formulation}
\begin{gathered}
E_{\mathbf{p} j}:=\sum_{\mathbf{p} \in \mathcal{N}_{\mathbf{p}}} w_{\mathbf{p}} \| I_j\left[\mathbf{p}^{\prime}\right]-I_{j, o}[\mathbf{p}^{\prime}]-\beta(I_i[\mathbf{p}]-I_{i, o}[\mathbf{p}^{\prime}])\|_{\gamma}
\\
I_{j, o}[\mathbf{p}^{\prime}] = e^{-\frac{t_{r,j-1}}{\tau_c}} \cdot I_{j-1}[\mathbf{p}^{\prime}] \\
I_{i, o}[\mathbf{p}^{\prime}] = e^{-\frac{t_{r,i-1}}{\tau_c}} \cdot I_{i-1}[\mathbf{p}^{\prime}] \\
% a_i=e^{-\frac{t_{e,i}}{\tau_h}} \\
% b_{i-1}=e^{-\frac{t_{r,i-1}}{\tau_c}}\\
% a_j=e^{-\frac{t_{e,j}}{\tau_h}}\\
% b_{j-1}=e^{-\frac{t_{r,j-1}}{\tau_c}}\\
\beta=\frac{1-e^{-\frac{t_{e,j}}{\tau_h}}}{1-e^{-\frac{t_{e,i}}{\tau_h}}}
\end{gathered}
\end{equation}
where $\mathcal{N}_{\mathbf{p}}$ is the set of pixels in the SSD, and $\|\cdot\|_{\gamma}$ is the Huber norm. In addition to using robust Huber penalties, a gradient-dependent weighting $w_{\mathbf{p}}$ \cite{engel2017direct} is applied.

%\subsection{SfM: Optimization}

%\aw{revisit position invariance discussion to indicate how this affects the optimizer}

To minimize the photometric error between the corresponding points in two frames, the optimizer then optimizes the heating and cooling time constants $\tau_c$ and $\tau_h$, instead of the brightness transfer variables in DSO. The optimization is accomplished using the Gauss-Newton algorithm in a sliding window \cite{leutenegger2015keyframe}. 

\section{Experiments and Results \label{sec:exp}}

In this section, the proposed photometric model for infrared sensors is evaluated on two datasets, the FLIR ADAS Dataset v1.3 \cite{FLIR_dataset} and the BU-TIV dataset \cite{wu2014thermal}. Both datasets contain RGB and thermal images for the same scene. The FLIR dataset contains a video sequence of cameras mounted on a vehicle moving in an area during nighttime while the BU-TIV dataset contains a video sequence of a daytime street scene recorded by stationary cameras. The experimental results are obtained on a 32-core Intel Xeon Silver 4110 CPU.

\subsection{Evaluations on FLIR Dataset}

\begin{figure}[h]
     \centering
     \begin{subfigure}[b]{0.22\textwidth}
         \centering
         \includegraphics[width=\textwidth, height=0.8\textwidth]{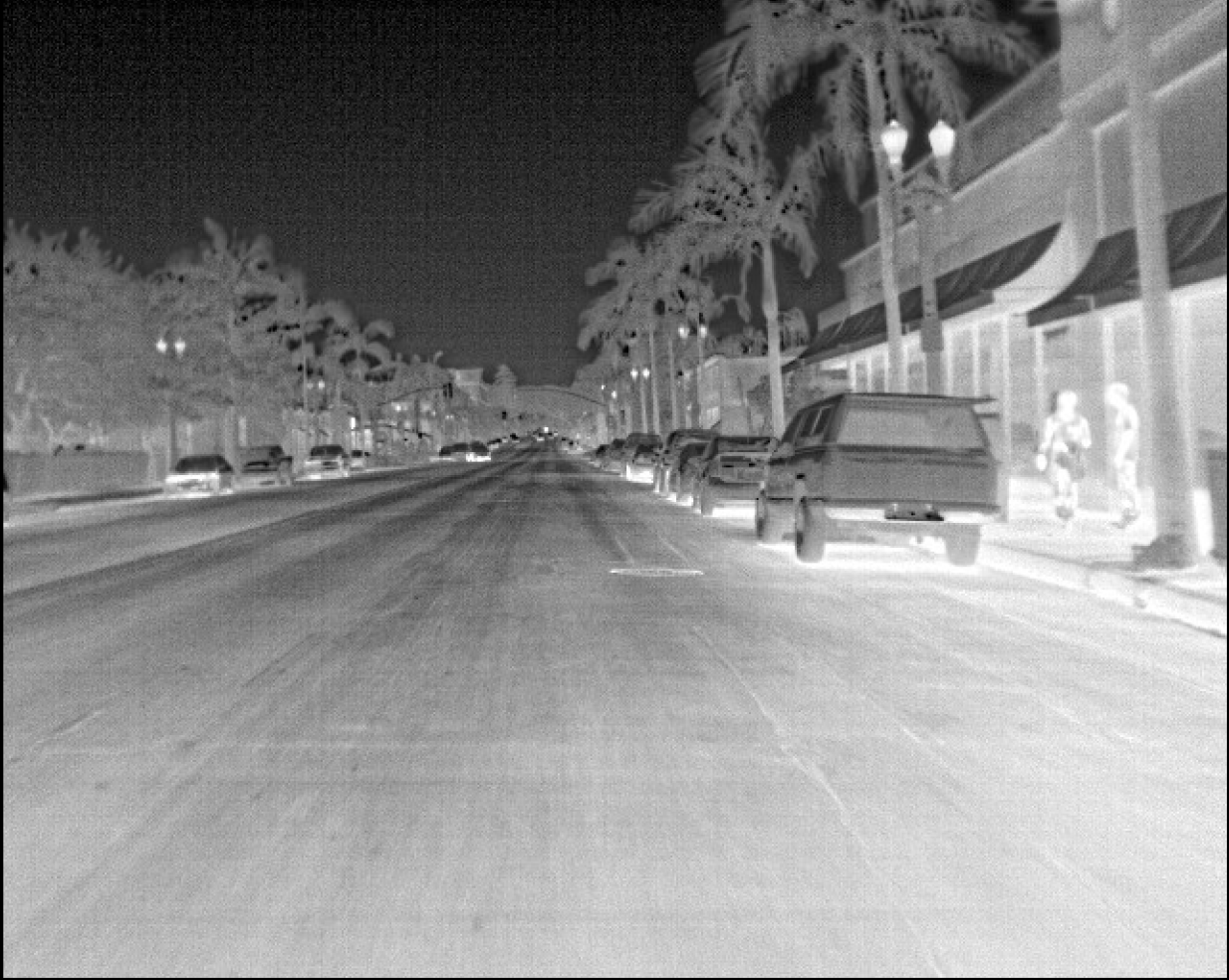}
     \end{subfigure}
     \hfill
     \begin{subfigure}[b]{0.22\textwidth}
         \centering
         \includegraphics[width=\textwidth, height=0.8\textwidth]{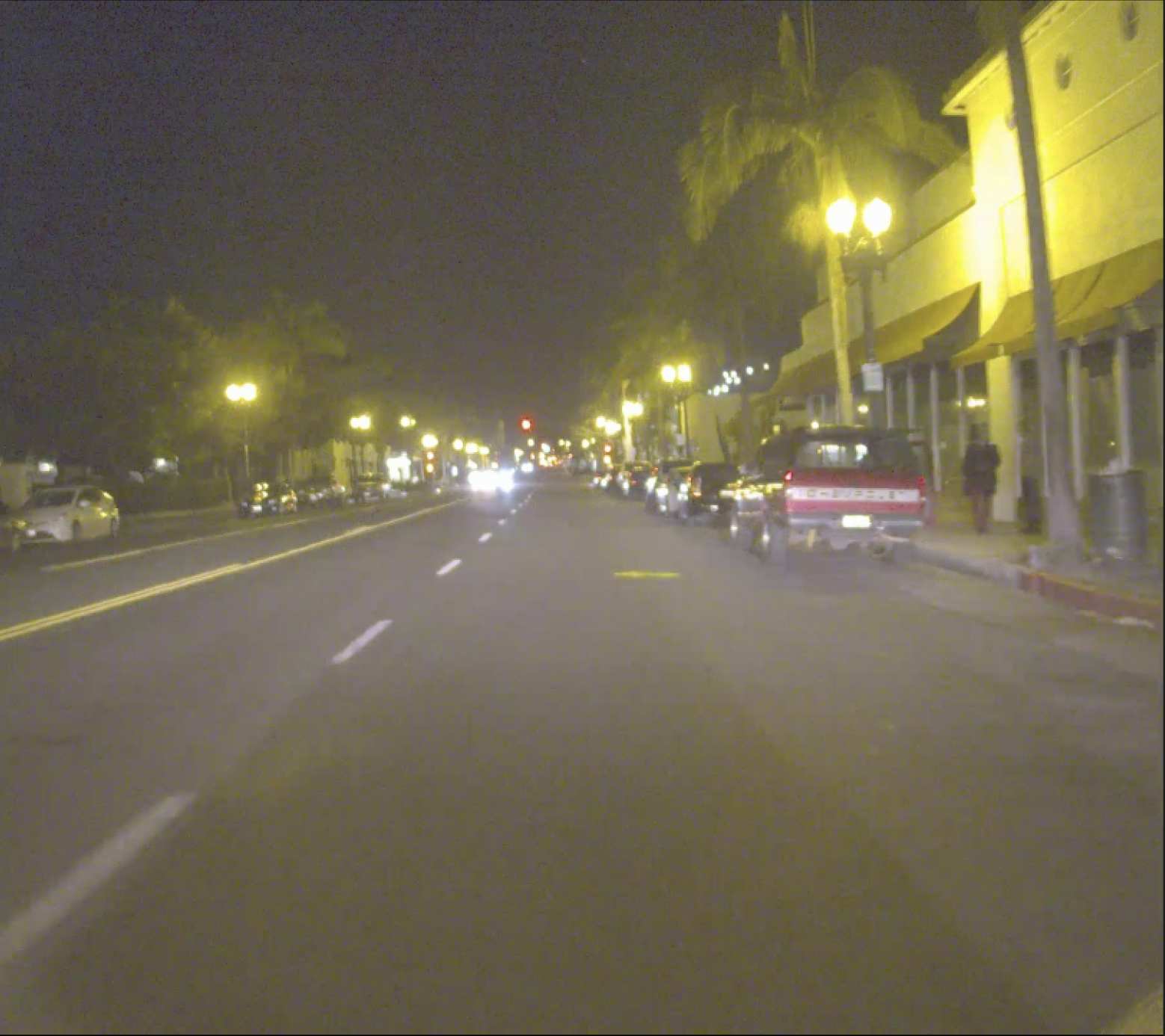}
     \end{subfigure}
     \hfill
     \caption{\label{fig:dataset_sample} An image pair sample from the FLIR ADAS dataset: IR image (left) and RGB image (right).}
\end{figure}

The FLIR ADAS Dataset consists of a video sequence of images taken from an IR camera and an RGB camera mounted to the front of a vehicle. The dataset was acquired with an RGB and IR camera mounted on a vehicle (car) where the IR sensor was a Teledyne FLIR Tau 2 thermal camera and the RGB a Teledyne FLIR Blackfly camera. Both RGB and IR videos were recorded at 30 frames per second (fps) under generally clear conditions during the night. Experiments use a subset of the complete ADAS dataset that corresponds to a video sequence of 600 images where the vehicle drives straight down a road at night. Example images from this video sequence are shown in Fig. \ref{fig:dataset_sample}. The results are summarized in two experiments:

\begin{itemize}
    \item Evaluations on the reconstruction quality of the road show that our photometric correction enables DSO to track more points on IR data and improve the accuracy of reconstruction.
    \item Evaluations on the camera motion show that with the proposed correction model, the trajectory is more stable and less deviated in terms of being at a certain distance away from the RGB camera trajectory.
\end{itemize}

%The RGB and IR cameras were housed in a single enclosure and were separated by a baseline distance of  48 mm within this enclosure. Thermal images were acquired with a FLIR Tau2 camera (13 mm f/1.0, 45-degree horizontal field of view (HFOV) and a vertical field of view (VFOV) of 37 degrees). RGB images were acquired with a FLIR BlackFly camera (4–8 mm f/1.4 16-megapixel lens with the field of view (FOV) set to match Tau2). 

%The FLIR ADAS Dataset v1.0.0 \cite{FLIR_dataset} is used for our experiments and evaluation. The FLIR Dataset provides a video sequence that is acquired via a thermal and an RGB camera mounted on a vehicle driving in the streets at night. The IR data is captured by a Teledyne FLIR \emph{Tau 2} thermal camera and the RGB by a Teledyne FLIR Blackfly camera at 30fps. The availability of both IR images and visible spectrum (RGB) is important for result comparison. Example images from the FLIR dataset are shown in Fig.\ref{fig:dataset_sample}.

\begin{figure}
    \centering
    \includegraphics[width=0.43\textwidth, height=0.27\textwidth]{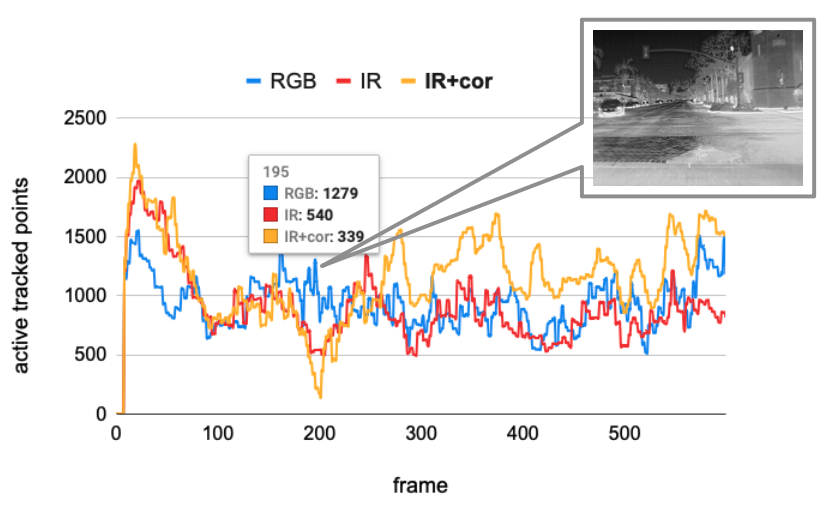}
    \caption{The number of tracked features is plotted for each video frame for ``\emph{RGB}" (blue), ``\emph{IR}" (red), and ``\emph{IR+cor}" (yellow) SfM estimates. The plot shows that the proposed IR photometric correction allows more stable feature detection which results in more tracked features leading to denser SfM estimates.}
    \label{fig:kf_points}
\end{figure}

\subsubsection{Evaluation Metrics}

Our photometric correction algorithm for IR pixel value correction was integrated into the code for the DSO algorithm \cite{engel2017direct} as a representation of a state-of-the-art SfM algorithm for RGB image sensors.  Experiments are performed using RGB and IR image data as input to the DSO algorithm. We consider 3 outputs: (1) the SfM estimates using the input RGB images, referred to as ``\emph{RGB}", (2) the SfM estimates using the input IR images without the proposed correction, referred to as ``\emph{IR}", and (3) the SfM estimates using the IR input images with the proposed correction, referred to as ``\emph{IR+cor}". 

\begin{figure*}%[htp]
\centering
\begin{tabular}{ccc}
  \includegraphics[width=.26\textwidth, height=.20\textwidth]{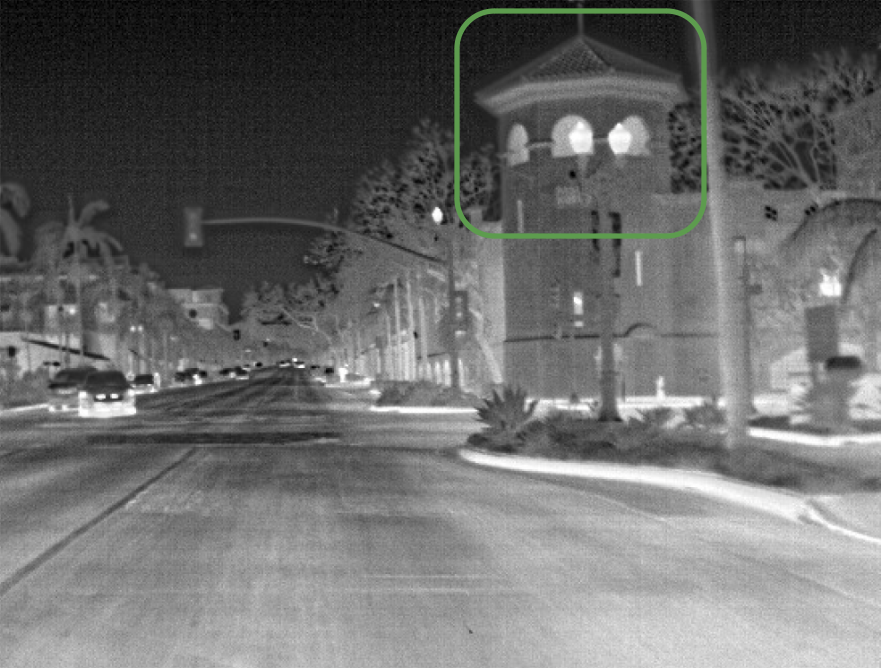}
  &
  \includegraphics[width=.26\textwidth, height=.20\textwidth]{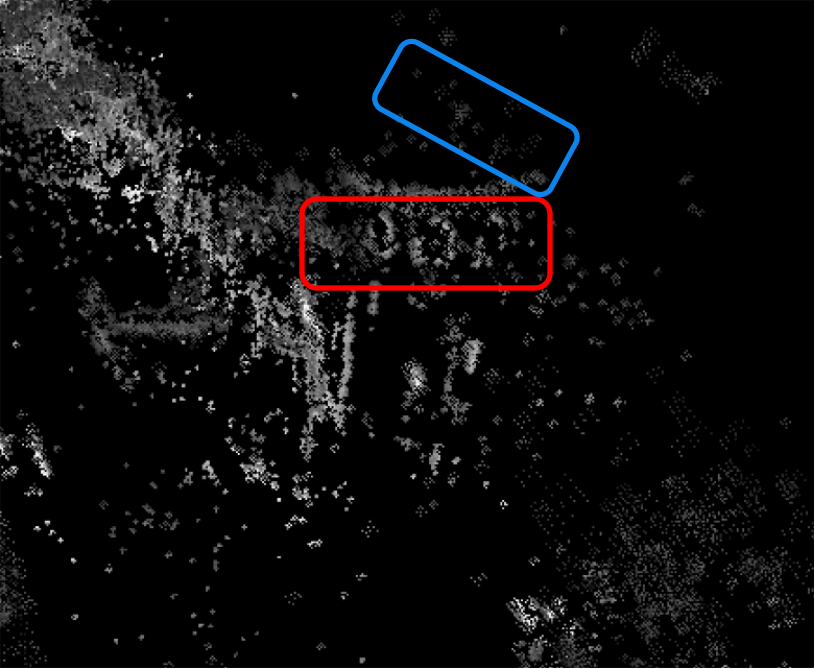}
  &
  \includegraphics[width=.26\textwidth, height=.20\textwidth]{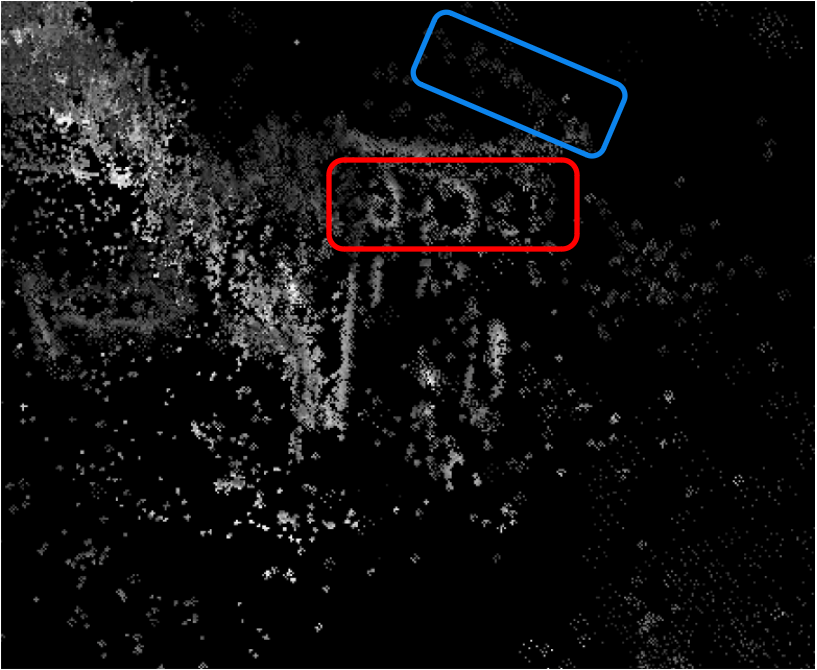}
  \\
    \textbf{(a)}  & \textbf{(b)} & \textbf{(c)}  
    \\[6pt]
  \includegraphics[width=.26\textwidth, height=.20\textwidth]{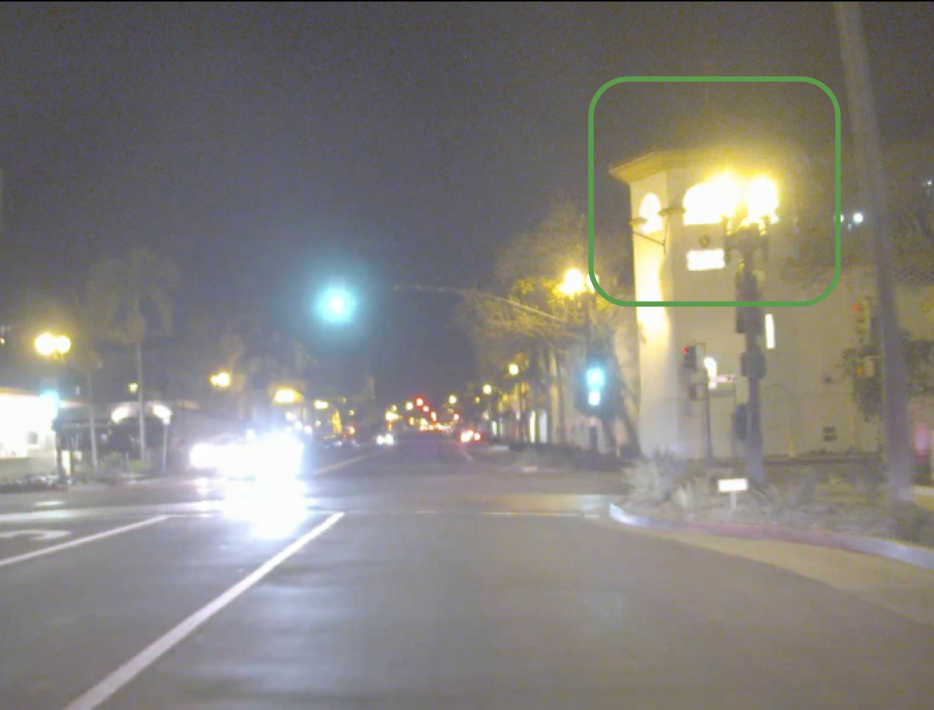}
  &
  \includegraphics[width=.26\textwidth, height=.20\textwidth]{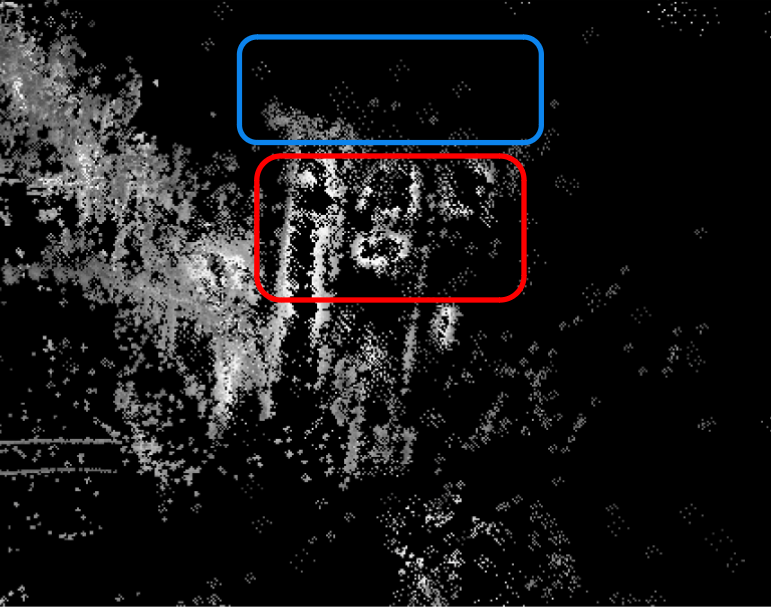}
  &
  \multicolumn{1}{b{.2\linewidth}}{}
  \\
  \textbf{(d)}  & \textbf{(e)}  
  \\[6pt]
\end{tabular}
\caption{\textbf{Qualitative examples.} (a) Infrared image from the FLIR ADAS Dataset. (b) ``\emph{IR}" point cloud reconstruction. (c) ``\emph{IR+cor}" point cloud reconstruction. (d) RGB image of the same scene from the dataset. (e) ``\emph{RGB}" point cloud reconstruction.}%\kb{use teh RGB, IR and IR+cor labels in the caption? you can still keep the full description, although it is in the text, but you could also drop them an just say RGB point cloud reconstruction, IR point cloud... , IR+cor point...  }}
\label{fig:quality}
\extralabel{fig:quality:a}{(a)}
\extralabel{fig:quality:b}{(b)}
\extralabel{fig:quality:c}{(c)}
\extralabel{fig:quality:d}{(d)}
\extralabel{fig:quality:e}{(e)}
\end{figure*}

As the only public infrared dataset containing certain data (a series of images taken from different viewpoints) that can be used for solving SfM problems, the FLIR ADAS dataset, however, has some limitations that pose challenges to our experiments: (1) the exposure time information of each frame is not available; (2) the 3D scan of the scene is not provided, and (3) the ground truth of the camera pose is not provided. Missing the accurate exposure time can undermine the advantage of the proposed photometric correction model for IR sensors. The lack of 3D scan and camera pose ground truth makes it difficult to analyze the exact improvements brought by the correction model.    

To overcome these drawbacks, based on the appearance of the objects shown in the dataset, for example, the street appears to be flat and the vehicle was driving toward one direction on the same lane, three hypothesizes are made for our experiments:

\begin{itemize}
    \item \textbf{Exposure Time Hypothesis:} The exposure time of each frame is around 10 milliseconds. 
    \item \textbf{Planar Road Hypothesis:} The road where the video was recorded can be approximated as a planar surface.\label{hypo:planar_street_hypo}
    \item \textbf{Straight-line Trajectory Hypothesis:} The trajectory of the camera, when the vehicle is driving straight on a road, can be approximated as a line.\label{hypo:straight_traj_hypo}
\end{itemize}

According to the FLIR Tau2 camera document \cite{tau2timeconstant}, the exposure time of the Tau2 camera is measured to be about 10 milliseconds. The value is then used as an approximation of the exposure time for each frame in the dataset. The Planar Road Hypothesis allows us to define a plane that fits the structure of the road to serve as a ground truth for the reconstruction quality evaluation. The Straight-line Trajectory Hypothesis provides an opportunity to evaluate the accuracy of the camera pose by looking into the deviation of the trajectory from the line.

%segment the road area and extract the points that fall on the road, then fit a plane to the point cloud as the road ground truth. The reconstruction quality is then analyzed on the reconstruction results from IR and RGB data respectively, by evaluating the distance of the road points to the road plane. 

%Specifically, the estimated camera trajectory is divided into 10 segments and under each segment, a rectangular area that contains the road area is defined. Based on the common sense of the height of a camera mounted in a car, such area is set to be 0.5$m$ underneath the trajectory segment, with a height of 0.25$m$. Additionally, since the vehicle in the dataset drives on the right, we define the width of the rectangular area to be 1$m$ to the right of the trajectory and 2$m$ to the left. 

\subsubsection{Structure Estimate Evaluation\label{sec:structure_estimate_evaluation}} 
%The point cloud of the street scene is generated using DSO respectively from the infrared (IR) data and the RGB data. In the experimental results presented in this section, we use ``RGB" to refer to the results from running native DSO on the RGB data with their RGB sensor photometric correction, with ``IR" being running native DSO on the IR data with the same RGB correction. The term ``IR+cor" is used for referring to running our modified DSO on the IR data with our thermal sensor photometric correction model mentioned in Sec. \ref{sec:ir_sensor_model}.

The Planar Road Hypothesis in Section \ref{hypo:planar_street_hypo} is used to evaluate the structure reconstruction accuracy. The ``ground truth" of the road is defined by: (1) segmenting the road area within the point cloud, (2) sampling from each point cloud the \textbf{same} amount of points located within the region around the road surface, and (3) merging the points from all three point clouds and fitting a plane to the combined road points. This way a common reference of the road surface is available for evaluation. %\kb{you have a lot more points that came from IR than from RGB. So it's not clear that is a fair comparison. What do these look like if you simply see how noisy the fit is within its own data set? Besides, who says the road is flat???  Same for the straight line? Note - it looks like RGB is noisier, but the sigma isn't really fair since you have far more IR and IR+cor features, and I would expect them to have MANY common features, yes?  If you can quickly look compared to only its own set and look at noise, that would be interesting, but you should down-play the bias and focus on noise (i think?) }

%\begin{table}[h!]
%\centering
%\begin{tabular}{c c c c c c}
% \hline
%  data &  kfs & total pts & road pts & RMSE & std\\
% \hline
% RGB & 219 & 55562 & 1395 & 0.0153 & 0.0088 \\
% IR & 181 & 42411 & 1304 & 0.0137 & 0.0073\\
% \textbf{IR+cor} & \textbf{246} & \textbf{65229} & \textbf{1890} & \textbf{0.0119} & \textbf{0.0065}\\
% \hline
%\end{tabular}
%\caption{Statistics of the road reconstruction performance. \kb{consider "rotating" the table to match table 2,3, etc in structure?}}
%\label{table:reconstruction_comp}
%\end{table}

\begin{table}[h!]
\centering
\begin{tabular}{c c c c}%p{2cm} p{0.7cm} p{0.6cm} p{0.6cm} p{0.6cm} p{0.6cm}}
 \hline
   & RGB & IR & \textbf{IR+cor} \\
 \hline
  kfs & 219 & 181 & \textbf{246} \\
  total pts & 55562 & 42411 & \textbf{65229} \\
  road pts & 1395 & 1304 & \textbf{1890} \\
  RMSE & 0.0153 & 0.0137 & \textbf{0.0119} \\
  std. & 0.0088 & 0.0073 & \textbf{0.0065} \\
 \hline
\end{tabular}
\caption{Statistics of the road reconstruction performance.}
\label{table:reconstruction_comp}
\end{table}

The ``kfs" and ``total pts" rows of Table \ref{table:reconstruction_comp} show that the proposed IR photometric correction model allows for more scene points to be tracked for the SfM estimation algorithm. The ``kfs" row denotes the total number of keyframes for the image sequence and the ``total" pts row indicates the total number of tracked features for the image sequence. As shown in the table, the SfM algorithm using the IR correction tracks 53.8\% more points (total pts) and 35.9\% more keyframes (kfs) than IR input images without photometric correction. Similarly, the SfM algorithm using IR correction tracks 17.4\% more points and 12.3\% more keyframes than RGB input images using the RGB image photometric correction. We also note that similar numbers are found for the point clouds identified as inside the segmented region around the road surface (road points). These results indicate that the IR photometric correction allows more points to be tracked and the resulting SfM solution, therefore, yields a denser 3D point cloud for both the camera motion trajectory (number of keyframes) and the scene structure.

The number of actively tracked features over the 600-frame video sequence is plotted in Fig. \ref{fig:kf_points}. Actively tracked points are used for both camera motion and scene structure estimation as each new frame in the video is measured. Overall running the ``\emph{IR+cor}" enables more points for tracking while both the ``\emph{IR}" and the ``\emph{RGB}" track fewer points but are comparable to each other. An interesting undulation of the curves shown in the figure occurs at frame index 200. This corresponds to a sequence of images within a large two-way street intersection. The RGB image sensor can track better in this particular context due to rich structural appearance data provided by the white lines and cross-walk textures on the ground which have little thermal contrast. This explains the decrease in tracked points for frames 150-220 from the IR image sensor. 

To evaluate the accuracy of the structure reconstruction identical sections of the estimated reconstruction data in the vicinity of the road surface were segmented from the complete structure estimate. In each case, the segmented surface points were compared against a pre-defined road plane. Evaluation of performance was done by computing the Root-Mean-Square deviation (RMSE) and the standard deviation (std) of the perpendicular distance between reconstructed 3D points and the road surface. As presented in Table \ref{table:reconstruction_comp}, ``\emph{IR+cor}" improves the reconstruction accuracy of the road by 15.1\% over the ``\emph{IR}" approach and 28.5\% over the ``\emph{RGB}" approach. The road points detected in ``\emph{IR+cor}" exhibit less noise relative to the road plane model having 10.9\% and 26.1\% less deviation than the results of the ``\emph{IR}" and ``\emph{RGB}" methods respectively. These results indicate that the proposed IR photometric correction model reduces the noise in the estimated scene structure.

\begin{figure}%[htp]
\includegraphics[width=.4\textwidth]{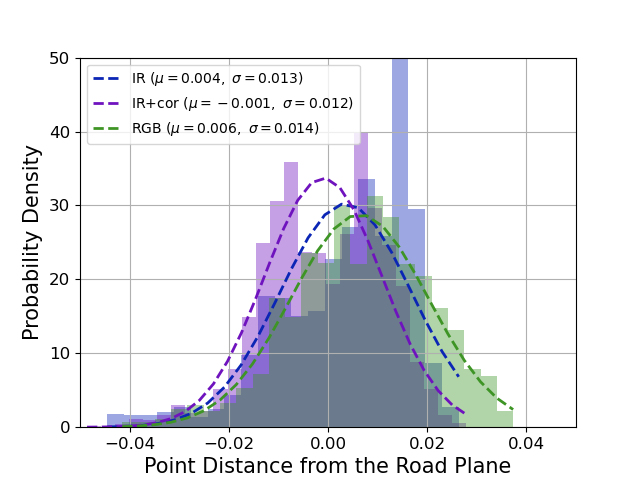}
\centering
\caption{Histogram and approximated Gaussian distributions of the fitting error of SfM-estimated 3D points to a common planar surface (the road plane). RGB (green) and IR (blue) show more error variance than the proposed IR photometric correction approach (purple).}
\label{fig:hist}
\end{figure}

This conclusion can be further supported by the histogram of the fitting error and error distribution in Fig. \ref{fig:hist}, from which we can also see that DSO on the IR data with infrared photometric correction achieves the best performance in terms of accuracy (closest to zero mean value) and stability (least standard deviation). Although RGB data can sometimes provide more features for detection and tracking (more tracked points), the reconstruction quality is not as well as the results from infrared data. This can be because the RGB intensity values are very prone to bad illumination conditions such as at night, which was the case when the dataset was recorded.

\subsubsection{Motion Estimate Evaluation} 

Performance analysis for the motion estimates uses a Straight-line Trajectory Hypothesis in Section \ref{hypo:straight_traj_hypo} for the vehicle motion and examines the observed errors of the estimated camera motion for each approach. In the video, the vehicle where the two cameras were mounted was driving on a straight street, as a result of which the trajectory of the cameras can be approximated to be straight as well. Towards this end, a 3D line segment is fit to the positions of the estimated camera trajectories for three approaches: (1) \emph{IR}, (2) \emph{IR+cor}, and (3) \emph{RGB}. Table \ref{table:trajs_comp} shows the RMSE (root-mean-square error) and standard deviation (std) of the perpendicular distances between camera positions and the estimated 3D line model. 

\begin{table}[h!]
\centering
\begin{tabular}{c c c c}
\hline
   & RGB & IR & \textbf{IR+cor}\\
\hline

 RMSE & 0.0128 & 0.0109 & \textbf{0.0106} \\
 std.   &   0.0061    & 0.0049 & \textbf{0.0045}\\
 \hline
\end{tabular}
\caption{Statistics of the trajectory estimation performance. }
\label{table:trajs_comp}
\end{table}

The RMSE error row of Table \ref{table:trajs_comp} indicates that the camera motion estimates for the infrared data using the proposed photometric correction model exhibit less error relative to the line model by a factor of 2.8\% for the IR method and 17.2\% for the RGB method. The standard deviation (std.) row of Table \ref{table:trajs_comp} indicates that the variability in the camera motion for the IR photometric correction is also less than that for the other two approaches. The reduction in noise for both IR estimates relative to the RGB data suggests that IR image data in this low-light context may provide advantages over RGB data for SfM estimation and that the proposed photometric correction further improves the SfM estimation performance in accuracy and stability.

\subsubsection{Qualitative Analysis}

The point cloud reconstructions from the IR and RGB data are illustrated in Fig. \ref{fig:quality} respectively. The green box in both Fig. \ref{fig:quality:a} and Fig \ref{fig:quality:d} includes the building of our interest. The red and blue boxes in the other three figures highlight the area where the reconstruction results differ. Compared to Fig. \ref{fig:quality:b},  Fig. \ref{fig:quality:c} shows sharper edges on the windows in the building (red box) and contains more points to reflect the edge of the top of the building (blue box). This indicates that the photometric correction model we propose can enable SfM algorithms to track more features (points) for reconstruction and the features are less noisy and more stable. As shown in Fig. \ref{fig:quality:e}, the top of the building (blue box) is completely missing and unrecognizable, and the point cloud of the windows in the building is very noisy (red box). This is because the top area of the building is interfered with by the illumination from the street light and RGB sensors can easily suffer from such illumination conditions and will fail to detect reliable features. The infrared sensors, however, are more robust in this scenario as they are sensitive to thermal contrast instead of photons. Note that the example image here comes from the same intersection area in the street as the image illustrated in Fig. \ref{fig:kf_points}, the point clouds here further prove that in this area of the street, more active points are tracked in the RGB data than the IR data due to the fact that rich structure but little thermal contrast are available in this area, as explained previously in Section \ref{sec:structure_estimate_evaluation}

\subsection{Evaluations on BU-TIV Dataset} 

To show that the proposed photometric correction model can also be applied to other IR image-processing contexts, the proposed photometric correction model is applied to the BU-TIV dataset for solving a human being tracking problem. The BU-TIV dataset is designated for the object-tracking problem using infrared data and consists of video sequences in different scenes that were recorded with FLIR SC8000 cameras. A subset of the dataset of a daytime crowded street view during a marathon competition was used. The results are summarized in two experiments: (1) Experiment 1 tracks pedestrians in IR image sequences. (2) Experiment 2 tracks the observed temperature for a stationary target in the IR video.

 \begin{figure}%[htp]
     \centering
     \begin{subfigure}[t]{0.2\textwidth}
         \centering
         \includegraphics[width=1.4\textwidth, height=0.9\textwidth]{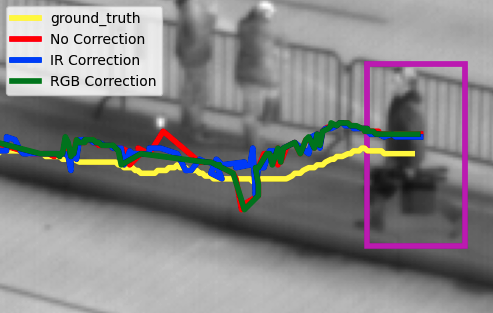}
         \caption{}    
     \end{subfigure}
     \hfill
     \begin{subfigure}[t]{0.2\textwidth}
         \centering
         \includegraphics[width=0.8\textwidth, height=0.9\textwidth]{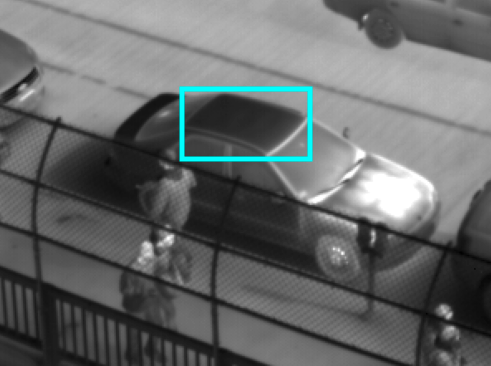}
         \caption{}
     \end{subfigure}
        \caption{Excerpts from the IR video in the BU-TIV dataset. (a) shows pedestrian (magenta box) tracking results relative to the ground truth (yellow), for the proposed correction (blue), RGB correction (green), and no correction (red). Table \ref{tab:trajs_comp} provides quantitative performance metrics. (b) shows the roof of the parked car (cyan box) where the temperature is tracked.}
        \label{fig:dataset_excerpts}
        \extralabel{fig:trajectories:a}{(a)}
        \extralabel{fig:trajectories:b}{(b)}
\end{figure}

\begin{table}[htp]
\begin{center}
\begin{tabular}{c c c c}
\hline
%\backslashbox[20mm]{Metric}{Method} & 
 & RGB &
IR &
\textbf{IR+cor} \\
\hline
%HD \cite{huttenlocher1993comparing}& 19.02 & 10.81 & 13.42 \\
%PCM\cite{witowski2012parameter} & 48.71 & \color{red}{45.33} & 46.98 \\
DF \cite{eiter1994computing} & 23.77 & 39.82 & \textbf{22.84} \\
DTW \cite{berndt1994using} & 4444.43  & 4339.27& \textbf{4257.07} \\
Mean Distance & 8.81 & 8.82 & \textbf{8.51}\\
Tracked frames count & 496 & 537  & \textbf{555} \\
\hline
\end{tabular}
\caption{Tracking differences measured by different algorithms between the estimated trajectories and ground truth.}
\label{tab:trajs_comp}
\end{center}
\end{table}

Experiment 1 evaluates the proposed photometric correction for pedestrian tracking from an IR image sequence. Results are shown in Fig. \ref{fig:trajectories:a} for three distinct photometric correction approaches: (1) RGB correction (\textit{RGB}), (2) no correction (\textit{IR}), and (3) the proposed correction (\textit{IR+cor}). Table \ref{tab:trajs_comp} shows the quality of each estimate as measured by three distinct metrics: (1) the Discrete Frechet (DF) distance \cite{eiter1994computing}
, (2) the Dynamic Time Warping (DTW) \cite{berndt1994using}, and (3) a custom metric referred to as the Mean Distance. Mean distance calculates the average distance between the person's estimated and actual positions for all corresponding frames. Table \ref{tab:trajs_comp} indicates that the proposed IR correction improves the performance across all three metrics, where lower scores are better. The last row shows the number of frames where the person is tracked and again the proposed IR correction algorithm outperforms the other cases.

%Following algorithms are commonly used by the literature for measuring the similarities between curves \cite{Jekel2019}: %Hausdorff distance (HD) \cite{huttenlocher1993comparing}, Partial Curve Mapping (PCM) \cite{witowski2012parameter}, 

Experiment 2 tracks the temperature of a parked car roof as shown in Fig. \ref{fig:trajectories:b} and seeks to analyze the stability of the temperature before and after applying the proposed IR photometric correction. Table \ref{tab:pixel_values} shows that the proposed photometric correction improves the stability of pixel intensity and, by extension, the estimate of the unknown constant temperature of the observed object.
%To further show that the proposed photometric correction for microbolometers sensors (IR sensors) brings stability of pixel intensity in the scene. The pixel value changes over frames (time) of the center point of the car roof (cyan box in Fig.\ref{fig:trajectories}) where the car is parked through the entire video is shown in Tab. \ref{tab:pixel_values}. Pixel values of the stationary car have the least variations when the proposed IR photometric correction is performed.

\begin{table}[h!t]
\begin{center}
\begin{tabular}{c c c c}
\hline
 & RGB &  IR & \textbf{IR+cor} \\
\hline
std. & 0.884 & 0.918 & \textbf{0.877} \\
\hline
\end{tabular}
\caption{Standard deviation of observed intensities for the roof of a parked car denoted as a cyan box in Fig. \ref{fig:trajectories:b} over time.}
\label{tab:pixel_values}
\end{center}
\end{table}

\section{Conclusion}

This article proposes a photometric correction model appropriate for microbolometer sensors typically integrated into infrared cameras. A new theoretical model for the pixel response is proposed and the parameters of this model are characterized. The photometric correction model was integrated into a state-of-the-art SfM algorithm where it was shown to improve upon the structure and camera motion estimates. Prior literature has made clear that photometric correction is an important component in improving the performance of SfM for RGB sensors and the results of this article indicate that appropriate models for infrared photometric correction also improve estimates in the infrared frequency regime. We hypothesize that the impact of the proposed infrared photometric correction further generalizes to potentially improve other computer vision algorithms when applied to IR image sensors, particularly those with the ``view invariance" or Lambertian assumption for the radiance of scene points over short motion distances.

%%%%%%%%% REFERENCES

\printbibliography
\end{document}